\documentclass{article}
\usepackage{amssymb,amsmath,epsfig,epstopdf,color,graphicx,subfigure}
\usepackage{epstopdf,array}
\newcolumntype{L}{>{\centering\arraybackslash}m{3cm}}
\usepackage{arxiv}

\usepackage[utf8]{inputenc} % allow utf-8 input
\usepackage[T1]{fontenc}    % use 8-bit T1 fonts
\usepackage{hyperref}       % hyperlinks
\usepackage{url}            % simple URL typesetting
\usepackage{booktabs}       % professional-quality tables
\usepackage{amsfonts}       % blackboard math symbols
\usepackage{nicefrac}       % compact symbols for 1/2, etc.
\usepackage{microtype}      % microtypography
\usepackage{placeins}
\usepackage{caption}

\title{Deep vs. Shallow Learning: A Benchmark Study in Low Magnitude Earthquake Detection}

\author{
  Akshat Goel\\
  Department of Computer Science\\
  University College London
  \And
  Denise Gorse\\
  Department of Computer Science\\
  University College London
}

\begin{document}
\maketitle

\begin{abstract}
While deep learning models have seen recent high uptake in the geosciences, and are appealing in their ability to learn from minimally processed input data, as 'black box' models they do not provide an easy means to understand how a decision is reached, which in safety-critical tasks especially can be
problematical. An alternative route is to use simpler, more transparent 'white box' models, in which task-specific feature construction replaces the more opaque feature discovery process performed automatically within deep learning models. Using data from the Groningen Gas Field in the Netherlands, we build on an existing logistic regression model by the addition of four further features discovered using elastic net driven data mining within the catch22 time series analysis package. We then evaluate the performance of the augmented logistic regression model relative to a deep (CNN) model, pre-trained on the Groningen data, on progressively increasing noise-to-signal ratios. We discover that, for each ratio, our logistic regression model correctly detects every earthquake, while the deep model fails to detect nearly 20 \% of seismic events, thus justifying at least a degree of caution in the application of deep models, especially to data with higher noise-to-signal ratios.
\end{abstract}

% keywords can be removed

\section{Introduction}
The geosciences have been transformed in recent years by a substantial growth in the quantity and quality of available data. This has spurred interest in machine learning methods for seismological tasks, and especially in deep learning models, as powerful feature extractors (see, for example, the benchmarking study of \cite{munchmeyer}). However, as 'black box' models, they do not provide an easy means to understand how a decision is reached. The aims of this paper are first to demonstrate the power of a 'white box' model and second to directly compare this model to a deep neural network.
Using data from the Groningen Gas Field in the Netherlands, we build on the logistic regression model of \cite{waheed} by the addition of four further features. We then evaluate the performance of the augmented model relative to the CNN of \cite{shaheen}, pre-trained on the Groningen data, on progressively increasing noise-to-signal ratios. We discover that, for each ratio, our logistic regression model correctly detects every earthquake, while the deep model fails to detect nearly 20 \% of
seismic events. It may thus be concluded that deep models should be treated with a degree of caution, since, in addition to their lack of transparency, they may not always perform ideally well.

\section{Data and Methods}

\subsection{Data acquisition, pre-processing, and partitioning}

The data used in this study, as in \cite{waheed} and \cite{shaheen}, were derived from the G-network, a network of 70 seismic stations set up by the Dutch government after public protest over increased induced seismicity due to gas extraction from the Groningen Gas Field. Each G-network
station has four geophones, between 50 m and 200 m, and we gather data from all four geophone levels as the comparison model of \cite{shaheen} requires this.
We downloaded seismograms for event examples from web services hosted by the Royal Netherlands Meteorological Institute (KNMI), applying the same date and magnitude ($\geq$ 0.2, resulting mean 1.05) selection criteria as in \cite{waheed}. This resulted in 2300 seismograms recording 47 events.
We were unable to obtain 4000 noise samples, as used in \cite{waheed}, from the 2017-18 period of the studies of \cite{waheed} and \cite{shaheen} and so used the same time of year in 2020-21. Additional negative examples for the increasing noise-to-signal phase of this study were drawn from this same period, with
38100 new samples collected. All waveform data were detrended, demeaned, and bandpass-filtered to frequencies of 5-25 Hz as in \cite{shaheen}. In addition, in order to be compatible with the CNN of \cite{shaheen}, the resulting seismograms were then downsampled by a factor of two.
We chose to segregate our train (60 \%), validation (20 \%), and test (20 \%) datasets by event rather than seismogram (seismogram-based segregation having been used in the studies of \cite{waheed} and \cite{shaheen}) in order to prevent data leakage from test to train datasets. Hence, for example, our test set may not contain
20 \% of the seismograms, as events are detected by varying numbers of G-network stations. This difference, and the differing time period from which noise data were derived, are not problematic for internal comparisons within this study. However, they should be borne in mind when comparing our
results directly with those of \cite{waheed} and \cite{shaheen}.

\subsection{Machine Learning Models}

The primary model used in this study is logistic regression (LR). The major appeal of this linear model is that its fitted weights can be used to understand the role and importance of different features. An
elastic net penalty 
\[
 [\alpha \sum_{i=1}^{p}\left|\beta_{i}\right|+(1-\alpha) \sum_{i=1}^{p} \beta_{i}^{2}]
\]
 where the $\beta_{i}$ are the (external inputs + 1) weights of the model, may be added to the LR loss function; for a suitable setting of the hyperparameter ${\alpha}$ this can encourage not only smaller weights but variable selection. In our work the elastic net penalty was used only for this latter purpose, as a means to filter new candidate features according to their importance, our final model being a simple, interpretable LR model as in \cite{waheed}.

We benchmarked against a convolutional neural network (CNN), a type of deep learning model originally designed for image data, and used in \cite{shaheen}. The CNN was not retrained as it had already been trained on data from the Groningen Gas Field. The CNN of \cite{shaheen} was designed to take advantage of the multiple geophone levels used in the G-network, leveraging the potential of the moveout pattern of energy to distinguish between disturbances originating underground (more likely to be seismic in origin) and ones originating at the surface (more likely to be noise).

\subsection{Performance measures}
In order to explore the behaviour of the models for increasing amounts of noise it was important to use a performance measure that is robust for imbalanced data. Accuracy is not a good choice in this respect, though we quote it as a secondary metric in our results as it is widely used elsewhere even for imbalanced datasets. Our preferred measure will be the more robust Matthews Correlation Coefficient, defined by
\[
MCC =	[TP*TN - FP*FN]/[[TP + FP]*[TP + FN]*[TN + FP]*[TN + FN]]^{1/2}
\]
where $TP$ denotes true positives, $TN$ true negatives, $FP$ false positives, and $FN$ false negatives. An $MCC$ of 1.0
corresponds to a perfect prediction, a value of 0.0 either random prediction or an assignment of all examples to a single class. It is this last behaviour that makes the $MCC$ valuable for imbalanced datasets, as the tendency of most machine learning models is to over-assign examples to the majority class.

\subsection{Use of the HCSTA and catch22 software packages}
Most of the work that has been done in earthquake detection with LR models has used seismologically derived features (see, for example, \cite{miranda}, and references therein). But it is also of interest
to explore statistically-derived features, as was done by \cite{waheed}, who used features from the HCTSA (Highly Comparative Time-Series Analysis) package of \cite{hctsa}. HCTSA offers around 7,700 candidate features, with the four used in \cite{waheed} being selected by a mixture of mechanisms integral to the package and qualitative judgement.
The catch22 MATLAB package \cite{catch-22} contains a subset of features the HCTSA authors found to be best-performing for time series classification. Because there are only 22 of these features it is easy to add them to the inputs of an LR model with an elastic net penalty and use this model's inherent feature selection ability to highlight potentially useful new features. This means of feature selection differs from that used in \cite{waheed} to choose the original four features, and it seemed
possible it could discover some potentially valuable features the selection method of the earlier paper had missed. In fact this was so, as will be shown in the Results section to follow.

\section{Results}
\subsection{Feature discovery using elastic net}
\FloatBarrier 
\begin{figure}[!ht]
\centering 
\includegraphics[scale=0.8]{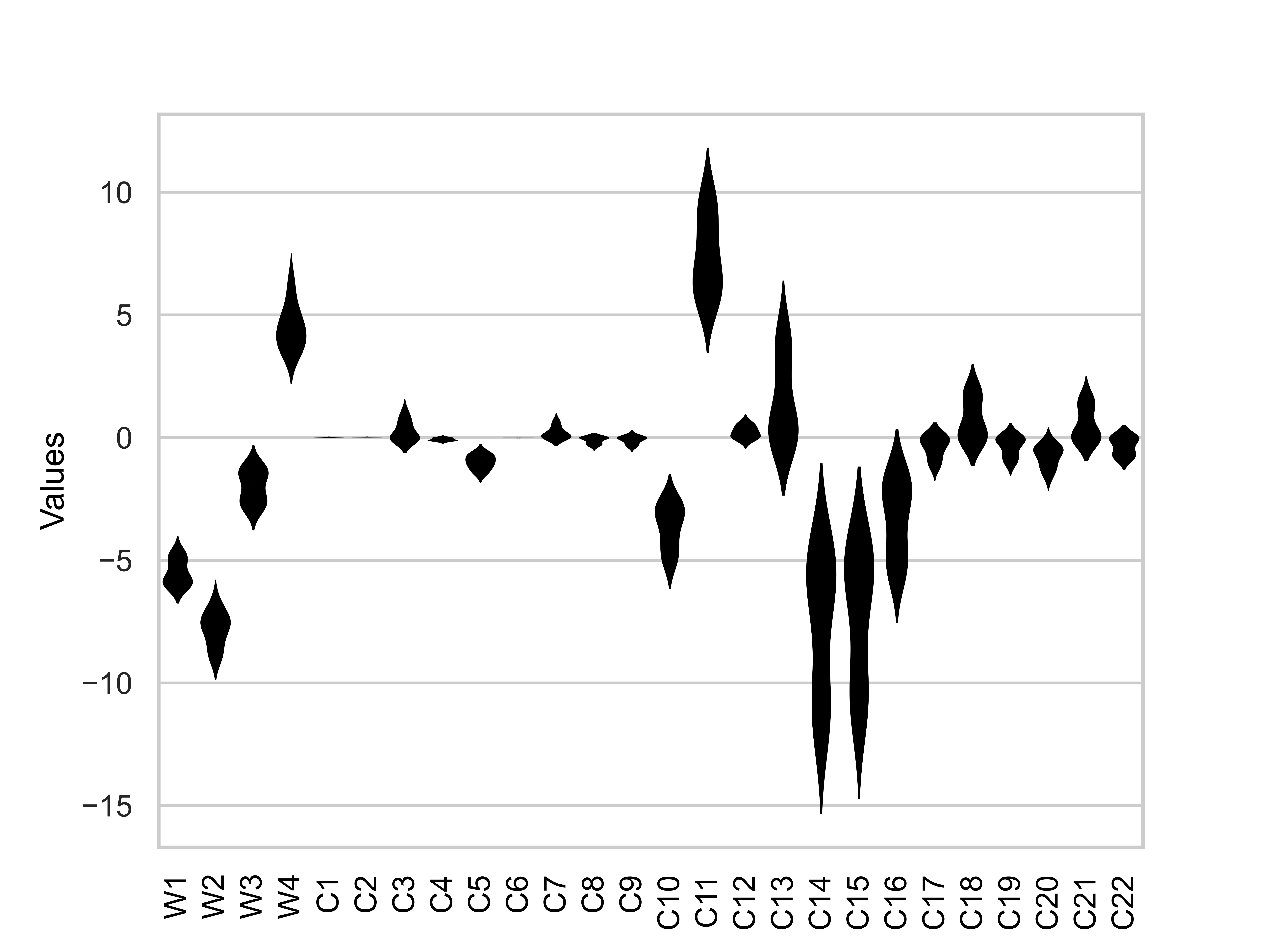} 
\caption{Distribution of feature values for the 72 best-performing elastic net models.}
\label{fig:g_config}
\end{figure}

200 runs of the elastic net model (which had 26 inputs, the original four HCTSA features from \cite{waheed} plus 22 new features from the catch22 package) were carried out. It was discovered that there was no single best set of weights; there were in fact many elastic net models with identical best
performance in terms of the validation MCC. The distribution of weights for each feature for these 72 equally well-performing models is shown in Figure 1, above. This figure confirms the value of the original four HCTSA features (W1 to W4), but also suggests the potential value of four catch22 features
(C10, C11, C14, and C15), with C11 especially promising. These four new features were added to the original four. The 18 less influential catch22 features and the use of the elastic net penalty were then discarded in order to have an interpretable LR model as in \cite{waheed}.

\subsection{Comparison of the logistic regression and CNN models at low noise-to-signal ratio}
Both the LR model of \cite{waheed} and our augmented LR model were trained at the noise-to-signal ratio of 1.73:1 used in \cite{waheed}. The CNN of \cite{shaheen}, which was not retrained by us, had been pre-trained at a ratio of 1.47:1. Both models were tested in this subsection at the ratio of 1.73:1 used for testing in \cite{waheed}. From Table 1 it can be seen the original LR model is able to achieve an MCC of 0.9691, but that our augmented model is even so able to statistically significantly improve on it. However, the CNN does substantially less well. From the confusion matrices in Table 2 it can be seen that the CNN is very effective (having no false positives) in classifying noise, but makes a substantial number of errors (false negatives) on earthquake data, failing to detect around 20 \% of the earthquake examples. As will be seen in the next subsection, the CNN's focus on correctly detecting noise becomes only more problematic as the proportion of noise to signal increases.

\FloatBarrier
\begin{table}[!ht]
\centering
\begin{tabular}{|l|l|l|l|l|l|}
\hline 
\textbf{Model} & \textbf{Model description} & \textbf{Model type} & \textbf{Test Acc.} & \textbf{Test MCC} & \textbf{Significant at 5 \%}\tabularnewline
\hline 
\textbf{1} & Model of \cite{waheed} & LR & 98.56 & 0.9691 & -\tabularnewline
\hline 
\textbf{2} & As above + selected catch22 & LR & 99.18 & 0.9824 & Yes (2 > 1) \tabularnewline
\hline 
\textbf{3} & Model of \cite{shaheen} & CNN & 82.94 & 0.8648 & Yes (3 < 2, 1)\tabularnewline
\hline 
\end{tabular}
\caption{\label{extend_test}Test performance at noise ratio 1.73:1 of the three models considered, in terms of both accuracy and MCC, where > (<) denotes that one model is statistically significantly better (worse) than another.}
\end{table}

\FloatBarrier
\begin{table}[!h]

\label{tab:cf_1}
\centering %
\begin{tabular}{|l|l|l|}
\hline 
 & Predicted negative & Predicted positive\tabularnewline
\hline 
True negative & 1313 & 5\tabularnewline
\hline 
True positive & 0 & 763\tabularnewline
\hline 
\end{tabular}
\caption*{Confusion Matrix: LR Model of \cite{waheed} + Selected catch22 Features}
\end{table}

\begin{table}[!h]

\centering %
\begin{tabular}{|l|l|l|}
\hline 
 & Predicted negative & Predicted positive\tabularnewline
\hline 
True negative & 1318 & 0\tabularnewline
\hline 
True positive & 127 & 627\tabularnewline
\hline 
\end{tabular}
\caption*{\label{cf_2}Confusion Matrix: CNN Model of \cite{shaheen}}
\caption{Confusion matrices for the better-performing of the two logistic regression models (model of \cite{waheed} + selected catch22) and for the benchmark CNN model of \cite{shaheen}.}
\end{table}

\subsection{Performance of the models on less balanced datasets}
Table 3 shows that as the noise ratio is increased all the models, which were trained on low ratios as described above, find classification more difficult. It can be seen, however, that the four catch22 features are of increasing value as the noise increases: at the highest noise-to-signal ratio considered in this study, 50:1, the augmented model's MCC is around 36 \% better than that of the original model.

\FloatBarrier
\begin{table}[!ht]
\centering
\begin{tabular}{|l|l|l|l|l|l|l|}
\hline 
\textbf{No.} & \textbf{Model description} & \textbf{1.73:1} & \textbf{5:1} & \textbf{10:1} & \textbf{25:1} & \textbf{50:1}\tabularnewline
\hline 
\textbf{1} & Model of \cite{waheed} & 0.9691 & 0.7663 & 0.4062 & 0.2388 & 0.1642\tabularnewline
\hline 
\textbf{2} & As above + selected catch22 & 0.9824 & 0.8203 & 0.515 & 0.3164 & 0.2227\tabularnewline
\hline 
\textbf{3} & Model of \cite{shaheen} & 0.8648 & 0.8921 & 0.8998 & 0.9045 & 0.9061\tabularnewline
\hline 
\end{tabular}
\caption{\label{extend_test}Test MCCs for the two LR models of \cite{waheed} and the benchmark CNN of \cite{shaheen} as the noise-to-signal ratio is increased from the baseline of 1.73:1 used in \cite{waheed}.}
\end{table}

It is not a surprise that the LR models find prediction on less balanced data harder. What is a surprise is that the CNN appears to find it easier. However, inspection of the confusion matrices for the higher noise ratios explains this: at these higher ratios the CNN makes the same errors on the earthquake
examples as it did at the starting ratio of 1.73:1, but continues to classify all noise examples perfectly; as the proportion of noise increases, the CNN therefore necessarily appears to do better.

\section{Conclusions}
This study benchmarked an eight-input logistic regression (LR) model (four new input features, discovered via the use of the catch22 software package combined with an elastic net feature selector, having been added to the model of \cite{waheed} against a convolutional neural network (CNN)
pre-trained on data from the same source, the Groningen Gas Field in the Netherlands \cite{shaheen}. Tests were carried out both at the low noise-to-signal ratio of \cite{waheed} and at higher noise ratios. While the CNN scored higher as this ratio increased, as measured by the Matthews Correlation Coefficient (MCC), the mistakes made by the LR model were notably all false positives, which can be reduced by seismological agencies using pre-existing or easily adoptable \cite{miranda} processes, while the mistakes made by the CNN were the more serious and problematic false negatives. We do not claim deep learning models have no place in seismology. They can be remarkably powerful; the EQTransformer model of \cite{earthquake-transformer} has been shown particularly effective in
recent benchmarking trials \cite{munchmeyer}. However, deep models need to be trained with care and have yet to be proven on substantially imbalanced datasets. A degree of caution may thus be warranted and the time may not yet be right for such models to be deployed in practice.

\section{Acknowledgments}
The authors would like to thank Umair bin Waheed for helpful discussions and advice, and Ahmed Shaheen for the provision of the pre-trained CNN model used in \cite{shaheen}.

\bibliographystyle{unsrt}  
\bibliography{references}

\end{document}